\documentclass[runningheads]{llncs}
\usepackage{graphicx}
\usepackage{xcolor}
\usepackage{kotex}

\usepackage{epsfig}
\usepackage{graphicx}
\usepackage{amsmath}
\usepackage{amssymb}

\definecolor{dark_cyan}{RGB}{0,139,139}

\newcommand{\hdh}[1]{\textcolor{black}{#1}}
\newcommand{\sam}[1]{\textcolor{black}{#1}}
\newcommand{\nj}[1]{\textcolor{black}{#1}}

\begin{document}
%
%\title{Contribution Title\thanks{Supported by organization x.}}

\title{The U-Net based GLOW \nj{for} Optical-Flow-free Video Interframe Generation}


\author{Saem Park \and %\orcidID{0000-0002-9727-4272} 
Donghoon Han \and %{\tt\small dhk1349@snu.ac.kr}
Nojun Kwak} %{\tt\small nojunk@snu.ac.kr}

%\author{First Author\orcidID{0000-0002-9727-4272} \and
%Second Author\inst{2,3}\orcidID{1111-2222-3333-4444} \and
%Third Author\inst{3}\orcidID{2222--3333-4444-5555}}
%
\authorrunning{Saem Park et al.}
\institute{Seoul National University, Seoul, Korea \\
{\tt\small (parksaem2,dhk1349,nojunk)@snu.ac.kr}
%\andLG Electronics
}
\maketitle              % typeset the header of the contribution
\begin{abstract}

Video frame interpolation is the task of creating an interframe between two adjacent frames along the time axis. So, instead of simply \iffalse adding\fi\hdh{averaging} two adjacent \hdh{frames} to create an intermediate image, this operation should maintain semantic continuity with the adjacent frames. \sam{\nj{Most} conventional \nj{methods use} optical flow, and various tools such as occlusion handling and object smoothing \nj{are} indispensable.} \sam{Since the use of these various tools leads to complex problems,} we tried to tackle the video interframe generation problem  \sam{without using problematic optical flow}. \hdh{To \nj{enable this},} \sam{we \nj{have tried} to use a deep neural network with an invertible structure, and} developed an U-Net based Generative Flow which is a modified normalizing flow. In addition, we propose a learning method with a new \nj{consistency loss in the latent space} to maintain semantic temporal consistency between frames. 
%Our method has the advantage of being able to form intermediate images by simply interpolating the outputs of the network without referencing the optical flow. 
%The resolution of the generated image is guaranteed to be identical to that of the original images by using an invertible network. 
%Furthermore, as it is not a random image like \nj{the ones by generative} models, \hdh{our network guarantees} stable outputs without flicker. 
% Through experiments, we \sam {confirmed the feasibility of the proposed algorithm and} \hdh{would like to suggest the U-Net based Generative Flow as a new possibility for baseline in video frame interpolation.}
\sam{This paper is meaningful in that it is the world's first attempt to use invertible networks instead of optical flows for video interpolation.}
\end{abstract}

%
%
%

%
%\noindent Displayed equations are centered and set on a separateline.
%\begin{equation}
%x + y = z
%\end{equation}
%Please try to avoid rasterized images for line-art diagrams and
%schemas. Whenever possible, use vector graphics instead (see
%Fig.~\ref{fig1}).

%\begin{figure}
%\includegraphics[width=\textwidth]{fig1.eps}
%\caption{A figure caption is always placed below the illustration.
%Please note that short captions are centered, while long ones are
%justified by the macro package automatically.} \label{fig1}
%\end{figure}

%%%%%%%%% BODY TEXT
\section{Introduction}
In the introduction, we want to explain the necessity of generating video interframes, existing methods of using deep neural networks, general optical flow problems, and \nj{the basics of} invertible networks.

%\subsection{The necessity of generating video interframes}
%Interframe generation, which creates an intermediate frame using information from the front and rear frames, is one of the techniques frequently used in TV systems.
%Most input video sources have a refresh rate of 24,30,60 Hz per second, but TV output can often output a refresh rate of 120 Hz per second.
%To do this, the TV interpolates over the temporal phase to form an intermediate image, producing a high frame-rate video of 120 Hz.
%Because simple synthesis of the front and rear frames cannot produce smooth video and judder occurs, the interframe generation algorithms generally move the front or rear frames in the motion vector direction using optical-flow.
%Optical-flow is information indicating in which direction each pixel of a frame is currently moving by using the correlation between the front and rear frames. Traditionally, a human-made program was used to calculate the optical-flow and use it to create an intermediate image.

\subsection{The necessity of generating video interframes}
Interframe generation, which creates an intermediate frame using the information \nj{of two consecutive frames}, is one of the techniques frequently used in TV systems.
%Most input video sources have a refresh rate of 24,30,60 Hz per second, but TV output can often output a refresh rate of 120 Hz per second.
\hdh{Most input video sources have a refresh rate of 24, 30, \nj{and} 60 Hz per second, but TV can often \nj{output frames} at a refresh rate of 120 Hz per second.}
To \nj{achieve} this, the TV interpolates over the temporal phase to form an intermediate image, producing a high frame-rate video of 120 Hz.
Because the simple synthesis of the front and rear frames cannot produce smooth video and \nj{the accompanying judder is inevitable}, the interframe generation algorithms generally move the front or rear frames in the motion vector direction using optical flow.
%Optical flow is information indicating in which direction each pixel of a frame is currently moving by using the correlation between the front and rear frames. Traditionally, a human-made program was used to calculate the optical flow and use it to create an intermediate image.
\hdh{Optical flow~\cite{barron1994performance,horn1981determining} is the information indicating in which direction each pixel is moving in a frame by using the correlation between the front and rear frames. Traditionally, a rule-based program was used to calculate the optical flow and \nj{algorithms use} it to create an intermediate image.}

\subsection{\sam{Video interpolation method using DNN}}
% The use of DNN technology for inter frame generation first appeared around 2017, and it was a method of interpolating the front and rear frames by learning a network to generate a supervised optical-flow like Flow-net\cite{Dosovitskiy_2015_ICCV},\cite{Ilg_2017_CVPR}.
% Super SlowMo\cite{Jiang_2018_CVPR} can create intermediate frames by learning the unsupervised optical flow using high-speed framed video shot with a high-speed camera.
\hdh{In the DNN field, several attempts \nj{have been} made to generate interframes \nj{since around} 2017. One attempt was interpolating the front and rear frames by training a network to generate a supervised optical flow like Flow-net~\cite{Dosovitskiy_2015_ICCV,Ilg_2017_CVPR} and PWC-Net~\cite{sun2018pwc}. The other attempt was Super SloMo~\cite{Jiang_2018_CVPR} which can create intermediate frames by learning the unsupervised optical flow using high-speed framed video shot with a high-speed camera.} 

In the former case, since learning is possible only with supervised information on optical flow, there is an inconvenience of making motion information for each video in a frame unit, which causes difficulties in generating training data. %This has the problem of limiting the set of training you can learn and add enormous costs to your training data.
\hdh{This is a costly job and \nj{the models} have to struggle with \nj{limited} training data.} 
In the latter case, only the front and rear frames are used as inputs, and the optical flow is predicted by itself without additional supervised optical flow information. This method does not require any additional information about the optical flow in the training data, so the training can be performed with any video input.
% In the Super \hdh{SloMo,}~\cite{Jiang_2018_CVPR} the network was trained mainly using \nj{slow-motion videos with a frame rate above 240Hz}, which has several advantages. 
% \hdh{First of all, high frame rated videos carry much more frames than normal videos in a \nj{fixed} time. It means that videos move smoother and carry more contextual information. Taking the advantage of high frame-rated videos, the model can boost performance in predicting the context between the frames.}
% Also, the video is relatively clearly divided into foreground and background, and
% \hdh{the majority of movements are centered on horizontal and vertical moves rather than complex motions.}

\begin{figure}
\begin{center}
    \begin{tabular}{cc}
    \includegraphics[trim=300 207 300 100, clip, width=0.42\linewidth]{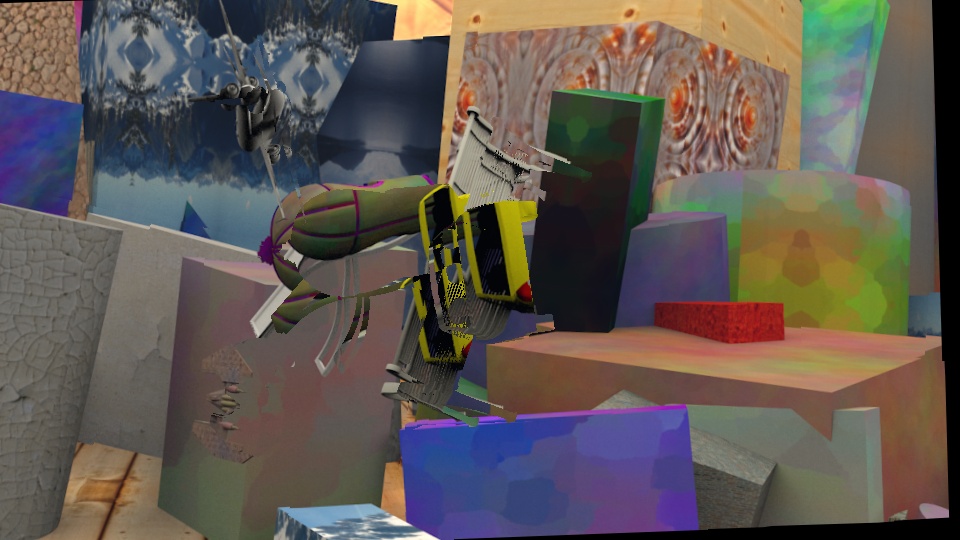} &
    \includegraphics[trim=40 50 200 286,clip,width=0.42\linewidth]{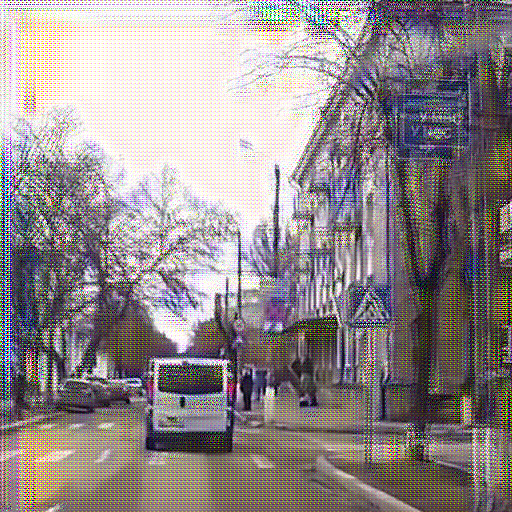}\\
    \footnotesize{(a) Limitation of optical flow} & \footnotesize{(b) Limitation of pixel shuffle} 
    %\\    Optical Flow & Pixel shuffle}
    \end{tabular}
\end{center}
\vspace{-5mm}
   \caption{\textbf{Limitations of previous works.} (a) shows \nj{a part of FlyingThings3D~\cite{MIFDB16} generated by warping the previous frame using} the ground truth optical flow. The object appears to be broken due to occlusion issues despite the ground truth being applied. (b) shows the interim results of learning with \nj{GLOW \cite{NIPS2018_8224}} \hdh{instead} of UGLOW. Pixel shuffle and 1x1 convolution cause side effects like the Bayer pattern.}
\label{fig:previous_problem}
\end{figure}

\subsection{Limitations of using optical flow}
As described above, most interframe algorithms use optical flow to move \nj{the} front and rear frames to create intermediate frames. However, due to optical flow's very endemic problems, the use of optical flow cannot produce a perfect interframe even when it refers to ground truth.  
Many methods use deep learning on video interframe generation~\cite{meyer2018phasenet,niklaus2017video,peleg2019net}. But \nj{they} cannot be free from optical flow problems.
\hdh{The problems of optical flow \nj{are} as follows}. The first is that it cannot accurately handle the boundaries of each object, and the second is that it is difficult to handle the occluded area.

The former boundary problem arises from the fact that when estimating the motion vector of an object, the boundary between the object and the background cannot be accurately known. In particular, if the boundary is predicted to be smaller than the object, it will be cut, causing serious problems. \hdh{For this reason, the motion vector of an object is generally set to erode the background more widely. However, if the background is eroded too much, it can cause side effects such as halo.} 
% Also, when calculating the optical flow of the entire image, \hdh{in some cases,} not all pixels are filled and \hdh{there may be empty holes.} The question arises as to how to interpolate these holes. In this case, when a simple \hdh{averaging} is used, a motion vector that does not exist in the actual video is generated, and when the surrounding value is expanded, the boundary becomes more visible. Another complex case is when there are multiple motion vectors within an object. 
% A common example is when an object rotates. In this \hdh{kind of circumstance}, the boundary of the motion is unclear and the object appears \hdh{to be} broken.

The latter occlusion is the processing problem of overlapping areas. This is a more complex problem than the former one. 
%This is because the correlation between the front and the back cannot be inferred and seen, so we have to find the correct answer for the domain that did not exist.
\hdh{It is because there is no correlation between \nj{the foreground and the background} when the generator tracks movements.} %The first process of treating occluded areas is to find out which one is ahead. Objects coming forward are taken from the next frame and the background should be deleted. 
As it is not easy to predict what will come \sam{forehead} when objects overlap each other, it is hard to predict the interframe.  %and even if the prediction is made, it is very difficult to create an intermediate image with only data in one temporal direction.
% \hdh{Even if the prediction is made, it is very difficult to create an intermediate image with only limited (temporal) directional data.}
%In addition to the two previously mentioned, there are issues such as not being able to find motion for a repetitive pattern or missing small objects, making it impossible to create a correct answer for optical flow unless it is a computer-graphics-made video. 
The occlusion problem has drawn much attention last year and has been dealt with by other scholars~\cite{Zhao_2020_CVPR}.
% \sam{The occlusion handling that covers the optical flow problem is very difficult \nj{but} important, and MaskFlownet~\cite{Zhao_2020_CVPR} which covers the problems of Flownet~\cite{dosovitskiy2015flownet,ilg2017flownet} \nj{drew many attentions last year.}}
%was chosen for oral presentation at CVPR 2020.}

% \hdh{Apart from several difficulties mentioned above, there are even more issues such as not being able to find motion in repetitive patterns or missing small objects. These drawbacks make it impossible for optical flow to generate a clear answer.}
% Looking at the image generated by applying the ground truth of the optical flow to the previous frame in Fig.~\ref{fig:previous_problem}(a), it is easy to understand why we \nj{do not} want to use it.
%In this paper, we propose a system that allows DNNs to achieve optimal results on their own without using optical flow. This is because, if necessary, the DNN itself generates appropriate information similar to optical flow.

% \hdh{In this paper, we propose the U-Net based Generative Flow that generates clean optical results. Without using optical flow, U-Net based Generative Flow is expressive enough to resemble the operations of optical flow and even outperforms the previous approaches.}

\subsection{Invertible Network}
To create an intermediate image without using an optical flow, we considered the use of an invertible network. 
In general DNN, due to the non-linear function and pooling process, it is impossible to perform reverse processing from the latent space to the original shape again. In this case, there is no way to recover the \hdh{original shape} from latent space due to data loss.

There are algorithms such as GLOW~\cite{NIPS2018_8224}, 
%MintNet~\cite{song2019mintnet},
BayesFlow~\cite{radev2020bayesflow}, and i-RevNet~\cite{jacobsen2018revnet,DBLP:journals/corr/abs-1802-07088} developed to prevent data loss in latent space in DNN. This bypasses half of the input channel and sends it to the next layer. This is used to generate the same convolution result as the forward path in the inverse path and restore the original shape using this. The idea of GLOW is that interpolation over a latent space can produce an intermediate image between two images that have semantic continuity. Our idea is that interpolating between two individual video frames similarly produces an intermediate frame with semantic continuity.

The invertible network has several advantages. The first point is that the image quality is not compromised and the original restoration is guaranteed so that the original quality can be maintained even for videos with sufficiently high input resolution. 
%The second advantage is that unlike generative models such as GAN, a 1:1 functional relationship between image and latency is established, ensuring \iffalse only one\fi \hdh{one and only} intermediate image and no random occurrence, enabling video processing that is susceptible to flicker.
\hdh{The second advantage is that unlike generative models such as GAN, \nj{\hdh{a}} 1:1 functional relationship between \nj{an} image and \nj{a point in a latent space} is established. This ensures the one and only intermediate image for each latent space \nj{without} random occurrence, enabling video processing without flickering.}

\subsection{Limitations of existing invertible network}
We want to interpolate between frames by applying an invertible network to \nj{a} video. %For this, some of the problems of existing invertible networks had to be solved. 
\hdh{For this, some of the problems of invertible networks had to be solved.}
%The biggest problem among them is that existing invertible networks such as \hdh{FLOW} and i-RevNet increase the channel while reducing the image size through pixel shuffle. 
\hdh{The biggest problem among them is that invertible networks such as FLOW and i-RevNet increase the \nj{number of channels} while reducing the image size through pixel shuffle. }This creates a fatal problem of generating Bayer patterns during interpolation, \sam{\nj{which} can be seen in Fig.~\ref{fig:previous_problem}(b).} We wanted the inter-pixel convolution process to look smooth enough for image processing, and for this, we had to remove the pixel shuffle. However, simply removing the pixel shuffle does not have a way to increase the \nj{number of channels} while maintaining the amount of information \nj{contained in} the input RGB channels, so the final network output also ends up with three channels. \hdh{In this case, generated information is also limited to three channels unless some measures like downsampling are \nj{considered}.}
Another problem is that FLOW uses 1x1 convolution to propagate information in half of the channel\hdh{s}, but it is difficult to divide the three channels of RGB in half. Also, since the 1x1 convolution does not refer to the surrounding pixels, only the data of the channel combined with the pixel shuffle is calculated. %In image processing, local convolution, which determines the type or shape of an object by referring to surrounding pixels, is advantageous, but the 1x1 convolution is not suitable for this. 
\hdh{In image processing, local convolution is desirable because it determines the type and shape of an object by referring to surrounding pixels, and the 1x1 convolution is not suitable for this kind of use. }
%In that i-RevNet also uses such a pixel shuffle, the desired effect did not come out sufficiently
\hdh{I-RevNet also uses pixel shuffle and its effect seemed vain.} \hdh{To sort out this issue, we have devised an U-Net based Generative Flow suitable for such image processing. }

\section{Contribution}
We introduce a DNN model suitable for generating video interframes using an invertible convolution network. The proposed network expands the number of channels without pixel shuffle and performs local convolution processing. Also, using U-Net, it is possible to refer to the upper \hdh{layer's} information in the form of a pyramid and \hdh{take} the advantage of being invertible.
Also, we propose a new learning method for video interpolation between frames. %This is unsupervised learning, and since the input image itself can be learned without optical flow, the network can generate the necessary information itself, and the intermediate image can be generated only through linear interpolation in a simple latent space.
\hdh{Since the network can learn the input image and generate the necessary information itself, the network can be trained in \nj{an} unsupervised manner. With the generated information, no data other than latent space is needed to generate the intermediate image through linear interpolation.}

The contributions of this paper are as follows:

\begin{enumerate}
%\item{Development of a new video frame interpolation method of generating intermediate images with simple linear interpolation in latent space without the use of optical flow.}
\item{\sam{\nj{The proposed method is} \hdh{the} world's first attempt} \hdh{ to suggest } \sam{a new approach} \hdh{for video frame interpolation} \sam{using an invertible deep neural network. Since no optical flow is used in our method, we can fundamentally avoid the problems} \hdh{that stem from optical flow.}}

\item{\nj{We propose} a \hdh{novel} U-Net based Generative Flow (UGLOW) that is invertible while utilizing local convolution and U-Net structure without pixel shuffle.}

%\item{Proving the effectiveness of the proposed method through \hdh{experiments}.}
\item{\sam{Using the proposed UGLOW, we} \hdh{also} \sam{propose a} \hdh{training} \sam{method that transforms continuous video frames into a linear latent space on the time axis.}}

\end{enumerate}

\section{Method}

\begin{figure*}
\begin{center}
\includegraphics[width=\linewidth]{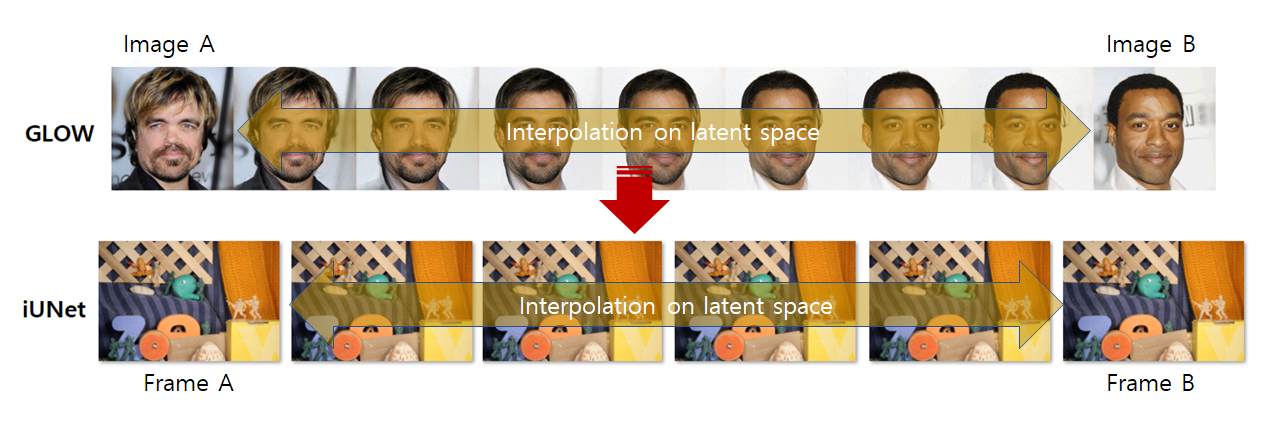}
\end{center}
\vspace{-8mm}
  \caption{\textbf{The concept of frame interpolation using UGLOW} The proposed UGLOW is a concept that interpolates video frames in latent space, like invertible networks such as GLOW.}
\label{fig:concept}
\end{figure*}

\subsection{Concept}
%In this section, we propose the idea concept and two major methodologies.
\hdh{In this \nj{section}, we propose the concept of U-Net based Generative Flow and its two major methodologies.}
%
%First, \iffalse the concept of the overall idea introduced.\fi 
\hdh{The idea of our concept is as follows.} 
%The easiest way to create \iffalse between frames\fi\hdh{intermediate frame} is to think of a linear interpolation between two \iffalse images\fi \hdh{adjacent frames}.
\hdh{When generating an intermediate frame, the easiest way would be the linear interpolation using two adjacent frames as \nj{inputs}.}
%However, since linear interpolation on simple images only produces double images that do not help with judder improvement, linear interpolation requires sending data into the transformed space where linear relationships between frames are established.
\hdh{However, such a method only produces \nj{overlay} images that do not help with judder. To generate a sophisticated intermediate frame, linear interpolation should be performed on \nj{a} transformed space after projecting adjacent frames onto \nj{it}. Then, a linear relationship with transformed space is established.}

%There are several conditions required for this.
\hdh{However, there are several conditions for such linear interpolation to perform as expected.} First, it should be possible to restore the original image from the converted space. Second, the transformed space should have a linear relationship with the time axis. %This is the condition  that\hdh{where the} invertible non-linear conversion is required. \iffalse We tried to find an idea to create a system that meets this condition. 
\hdh{To meet those conditions, the model should guarantee invertible non-linear conversion between the frame and \nj{the} converted space.}

The first condition got its idea from \nj{GLOW \cite{NIPS2018_8224}}. 
As you can see in Fig.~\ref{fig:concept}, 
GLOW \hdh{smooths} out interpolations between the two faces. We can think of \hdh{it as an} interpolation between two video frames. An invertible network like GLOW was an ideal \hdh{candidate for intermediate frame generation} because the latent space has a non-linear relationship with the input image. On top of that, the original image can be restored from the latent space.
%The second condition was a problem to be solved by learning so that the interpolation of the front and back frames in the latent space becomes latent of the intermediate frame.
\hdh{The second condition can be met if a model can generate the second frame's latent space with the interpolation of latent spaces from the first and third frame.}

We suggest the two novel methods to meet the concept and conditions.
The first is the \hdh{U-Net based Generative Flow (UGLOW)\nj{, which} is specially designed to learn video information in the most advantageous manner.} \hdh{In \nj{the} latter part,} we introduce the \hdh{very original} sub-modules that make up the network, how it is made to be invertible, and how the whole network is structured. \hdh{Details of UGLOW are \nj{described} in \nj{section} 3.2. }%UGLOW is guided in more detail in section 3.2.
%The second is how the learning was conducted using the UGLOW.
\hdh{The second is the loss metric that \nj{enables} the UGLOW to generate plausible frames without \nj{c}omplex algorithms.} Here we would like to introduce the idea used to make the latent space linear on the time axis. The detailed learning method will be introduced in \nj{sections} 3.3 and 3.4.

\begin{figure}
\begin{center}
    \begin{tabular}{cc}
    \includegraphics[width=0.45\linewidth]{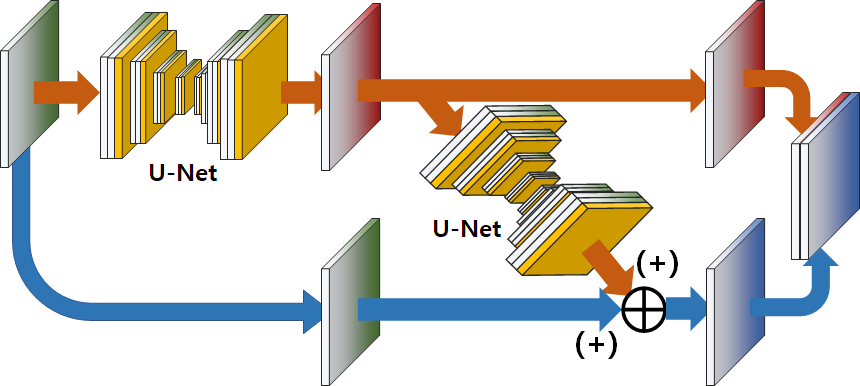} &
    \includegraphics[width=0.45\linewidth]{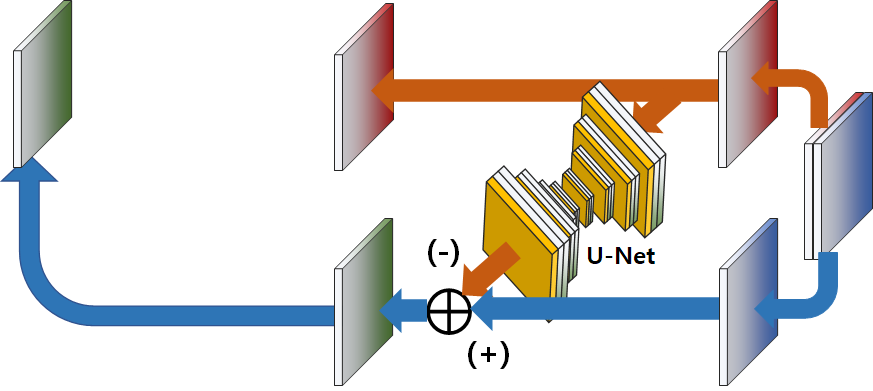}\\
    \footnotesize{\textbf{(a-1)}Channel Extend Module(Forward)} & \footnotesize{\textbf{(a-2)}Channel Extend Module(Backward)}\\
    \\
    \includegraphics[width=0.45\linewidth]{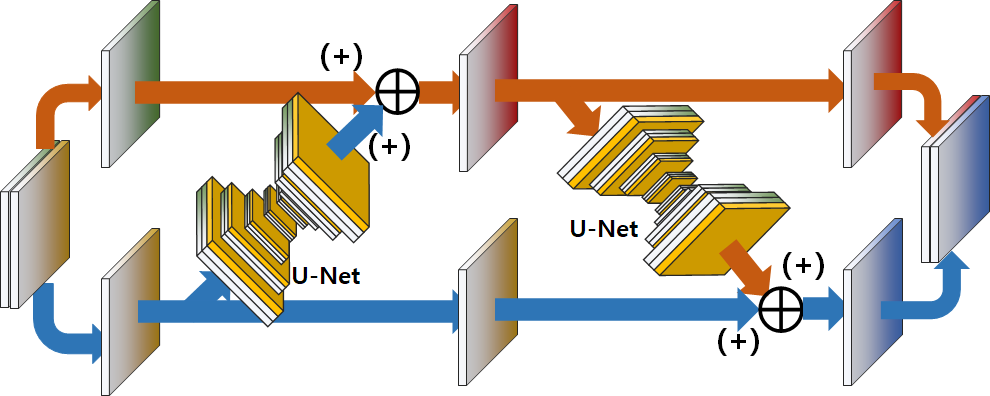} &
    \includegraphics[width=0.45\linewidth]{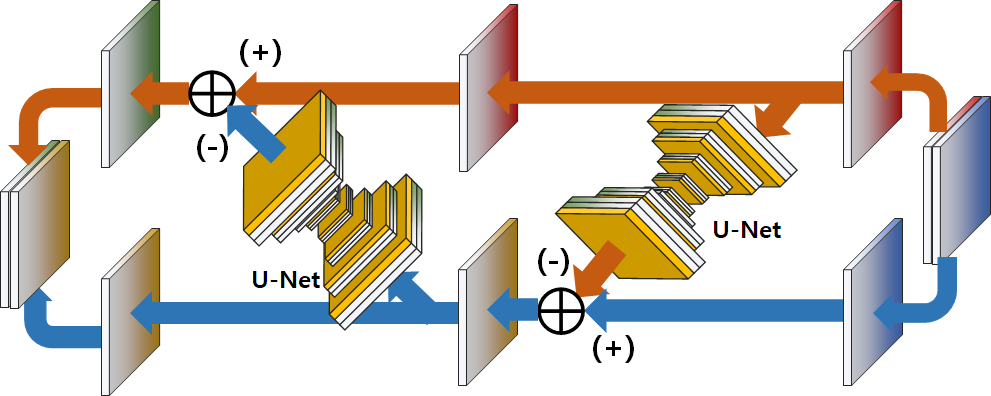} \\
    \footnotesize{\textbf{(b-1)}Channel Maintain Module(Forward)} & \footnotesize{\textbf{(b-2)}Channel Maintain Module(Backward)}\\
    \end{tabular}
\end{center}

\caption{\textbf{U-Net based Generative Flow Module.} These are invertible module using U-Net. Two types are depending on whether the channel is amplified or not, each showing how to recover the input from the output.}
\label{fig:UGLOW_block}
\end{figure}

\subsection{U-Net based Generative Flow (UGLOW) }
%We needed to devise an invertible network that can \iffalse analyze\fi\hdh{grasp} the context \iffalse of\fi\hdh{between the} frames to replace optical flow.
\hdh{We needed to devise an invertible network that can grasp the context between the frames to replace optical flow.}
% The range to be analyzed can range from the end of the screen to the other, depending on the degree of movement of the object. In this case, it is expensive to cover the entire area of information with convolution with a larger kernel size, so we found an efficient network that can reduce this.
\hdh{In some case\nj{s}, \nj{a} generator may have to track a large number of pixels depending on the size and movement of an object in the frame. If this is the case, it \nj{would be} expensive to cover the whole area with convolution \nj{with a large kernel size}. 
To address this issue, we tried to use \nj{an} efficient network and U-Net was \nj{a} feasible option.}
%U-Net is a network that allows you to view a wider range of upper and detailed information at the same time because subsampling occurs each time you go up to each layer.
\hdh{On the contracting path of U-Net, as the network propagates forward, subsampling enables the network to track a wider area with the same sized kernel. To be exact, each layer's subsampling enables kernel to refer twice a larger area. At the same time, skip connection is used to transmit information from the contracting path to the expansive path without loss, which prevents information loss due to the size of the bottleneck in the middle of the U-Net.}
Even in optical flow algorithms, pyramid structures with each layer halved in size are often used to detect large objects or fast motion vectors. \hdh{Therefore, the U-Net can be said to be \nj{a} suitable structure for \nj{an} intermediate frame generation task.}
\hdh{In addition, we developed the idea of skip connection and made the network invertible.}%Here, if we make good use of skip-connection, we can make it invertible.
%As mentioned in the introduction, \iffalse there was a problem that the existing invertible network uses pixel shuffle and 1x1 convolution,  3x3 or 5x5 convolution was used to refer to local information and processed without pixel shuffle.\fi\hdh{invertible network uses pixel shuffle and 1x1 convolution which can aggravate the output. U-Net based Generative Flow uses 3x3 or 5x5 convolution to refer to local information and processed without pixel shuffle. }The number of channels is preserved by removing the pixel shuffle, so we need to expand the output channel to increase the information. For this reason, it was necessary to make two types of blocks in which the channels are extended or maintained according to the number of input and output channels. We propose two \hdh{sub-block that modified U-Net to invertible form. }\iffalse local block structures in which U-Net is modified in an invertible form.\fi

\hdh{As mentioned in the introduction, the invertible network uses pixel shuffle and 1x1 convolution which can aggravate the output. U-Net based Generative Flow uses 3x3 or 5x5 convolution to refer to local information and \nj{works} without pixel shuffle. } 
\sam{Every U-Net consists of 4 down-blocks, 1 mid-block, and 4 up-blocks, each consisting of 2 convolution layers and Leaky-ReLU. The last output of the U-Net uses sigmoid as an activation so that it could adjust the image input normalized to 0-1.}
The number of channels is preserved by removing the pixel shuffle, so we need to expand the output channel to increase the information. For this reason, it was necessary to make two types of blocks in which the channels are extended or maintained according to the number of input and output channels. \hdh{We propose two sub-block\nj{s} that modified U-Net to \nj{an} invertible form.}

The first is a channel expend block in which the output has a channel twice as large as the input. This was designed to increase the number of channels as it is not enough to generate various information with only \nj{3-\hdh{channel}} input. (a-1) and (a-2) in Fig.\ref{fig:UGLOW_block} correspond to this. Since \hdh{UGLOW} is an invertible structure, of course, when reversed, the channel is reduced by half. Looking at (a-1) in Fig.\ref{fig:UGLOW_block}, it can be divided into the first half and the second half. 
The first half is to double the channel by attaching the original to the output of the U-Net like a skip connection, and it can be reversed by using the skip connected data as it is. %\iffalse In the second half, the data from the other side is added via U-Net, so that\fi
\hdh{In the second half, the data from the other side is added via U-Net, so} \sam{all the inputs are adjusted} \hdh{to} a non-linear transformation. The output created in this way can be restored by \hdh{(a-2)}. Since the output of the U-Net in the second half has the same value in the forward and reverse path, the input can be restored by \nj{simple subtraction instead of addition}. 

\begin{figure}
\begin{center}
\vspace{-3mm}
\includegraphics[width=0.95\linewidth]{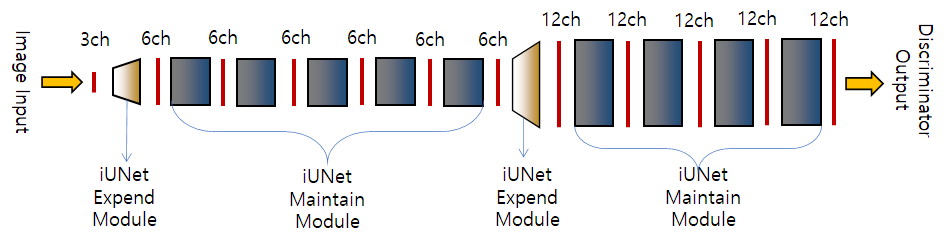}
\end{center}
\vspace{-7mm}
   \caption{\textbf{U-Net based Generative Flow architecture.} This is the overall structure of the UGLOW used in this paper.}
\label{fig:UGLOW}
\end{figure}

The second is a channel maintain block with the same number of channels as outputs and inputs, as shown in Fig.\ref{fig:UGLOW_block} (b-1) and (b-2). 
%This was created because of the limitation that the channel expansion block alone would increase the size of the channel exponentially and the thickness of the layer could not be sufficiently stacked due to cost issues.
\hdh{This block was designed to retain the number of channels. By using both extend block and maintain block, layers can be stacked deep enough while having enough channels.} Similar to the existing FLOW, this module uses half of the channel to process the other half and performs it in the opposite direction again, making it invertible while maintaining the number of channels.
Each \hdh{U-Net} has a quad down-block, a middle-block, and a quad up-block inside. As shown in Fig.\ref{fig:UGLOW}, %we made a deep neural network sufficiently composed of more than 100 layers of convolution by stacking 10 or more blocks by applying these two modules in combination.
\hdh{we designed a deep neural network that has more than 100 convolution layers by stacking \sam{11 invertible} modules.}
We name it UGLOW, meaning U-Net based Generative Flow.

\begin{figure}
\begin{center}
    \begin{tabular}{cc}
    \includegraphics[width=0.45\linewidth]{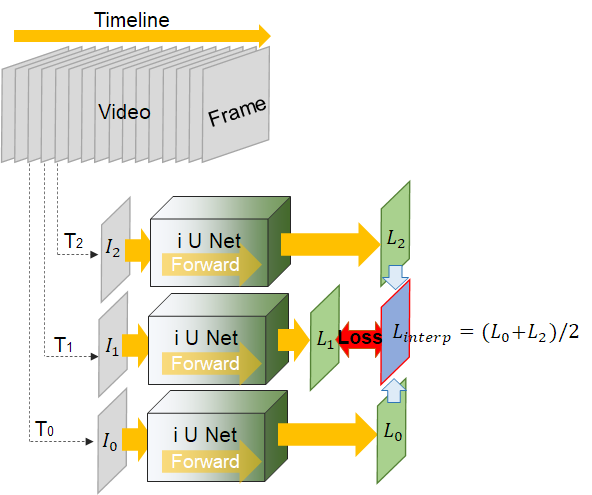}&
    \includegraphics[width=0.45\linewidth]{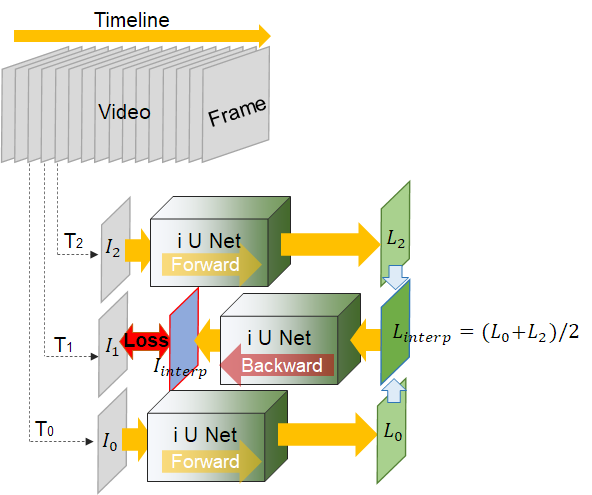}\\
    \textbf{(5-1)} Loss L & \textbf{(5-2)} Loss I
    \end{tabular}
\end{center}
    \vspace{-5mm}
   %\caption{\textbf{Distribution Loss Concept.} It makes distribution space linearity.}
   \caption{\textbf{Concept of the loss} Loss L makes latent spaces from consecutive frames have a linear relationship. Loss I makes reverse generated interframe as original like.}
\label{fig:loss_l}
\vspace{-2mm}
\end{figure}

\vspace{-7mm}

\subsection{Definition of Loss for Learning}

First, prepare a total of 3 consecutive frames from any video. These frames become training data, and even with a small number of videos, you can get a large number of training data with frame sliding.
Network learning proceeds with the simplest ideas.
\hdh{The core of the proposed learning method is to \nj{make sure that the three consecutive frames, whose center frame is generated from the other two} through \hdh{UGLOW}, have a linear relationship.} We suggest that simple linear blending in the latent space created through this learning method can represent the intermediate frame well in the time axis between the two frames.

Specifically, the input data of three consecutive frames are named $I_0$, $I_1$, $I_2$ and each image is transformed into a latent space via a reversible network. The latent spaces created in this way are named $L_0$, $L_1$, $L_2$. %At this time, we want to pay attention to the relationship between $L_2$ and $L_{inter}$, which created by linear interpolation of $L_1$ and $L_3$.
\hdh{At this time, \nj{the loss} is defined to minimize the difference between  $L_1$ and $L_{inter}$ which \nj{is} created by \nj{linearly interpolating} $L_0$ and $L_2$.}
The UGLOW we used can restore $I$ from $L$ through \nj{an} inver\nj{se} processing, and there is non-linearity between $I$ and $L$. \sam{This enables \nj{our} network to minimize $||\textrm{model.reverse}(L_{inter}) - I_1||^2$ and $||L_{inter} - L_1||^2$ at the same time.}

In our algorithm, the loss %can be defined in two main ways.
\hdh{ metric is designed to optimize two tasks.}
\begin{enumerate}
    \item {\hdh{The loss in the latent space aims to minimize the difference between the result of interpolation on the latent space and the latent space created from the intermediate frame: $Loss_{L}$.} When enough learning is \nj{done}, the latent spaces show a linear relationship with each other on the time axis. \sam{(Fig.\ref{fig:loss_l})}}
    \item {\hdh{The loss in the input space aims to \nj{minimize} the difference between the image $I_{inter}$ restored from $L_{inter}$ and \nj{$I_{1}$}: $Loss_{I}$. When enough learning is \nj{done}, the reversed intermediate \nj{frame} will match the actual intermediate frame.} \sam{(Fig.\ref{fig:loss_l})}}
\end{enumerate}

The detailed formula for the discriminator loss can be defined as follows.
\vspace{-2mm}
\begin{equation}
\textrm{model} = \textrm{UGLOW}(I)
\vspace{-5mm}
\end{equation}

\begin{equation}
\begin{aligned}
L_{0} = \textrm{model}(I_{0}) \\
L_{1} = \textrm{model}(I_{1}) \\
L_{2} = \textrm{model}(I_{2}) 
\end{aligned}
\vspace{-2mm}
\end{equation}

\hdh{
\begin{equation}
L_{inter} = ( L_{0} + L_{2} ) / 2
\vspace{-5mm}
\end{equation}
}
\hdh{
\begin{equation}
Loss_{L} = || L_{inter} - L_{1} || ^2.
\end{equation}
}
%\vspace{3mm}
The detailed formula for the reconstruction loss can be defined as follows.
\vspace{-2mm}
\begin{equation}
I_{inter} = \textrm{model.reverse}(L_{inter})
\vspace{-2mm}
\end{equation}
\begin{equation}
Loss_{I} = || I_{inter} - I_{1} ||^2.
\end{equation}
The final loss can be defined as follows.
\vspace{-2mm}
\begin{equation}
\vspace{-4mm}
Loss = w_L \times Loss_L + w_I \times Loss_I,
\vspace{-2mm}
\end{equation}
\nj{where} $w_L$ and $w_I$ are values for weight adjustment for each loss. 
When learning \hdh{UGLOW} by combining these two loss\hdh{es}, \hdh{UGLOW} learns how to set up a continuous frame to be a linear relationship in the latent space.
Due to the linear relationship of \hdh{latent spaces on the time axis,} simple blending can produce the result of an arbitrary mid-point without optical flow%\sam{(Fig.\ref{fig:recon})}
. Also, by ensuring inverse restoration, the image produced by the interpolated discriminator guarantees the same quality as the actual intermediate frame. 

%\begin{figure}
% \begin{center}
% \includegraphics[width=0.6\linewidth]{png/generation.png}
% \end{center}
%    \vspace{-3mm}
%   \caption{\textbf{How to make interframe.} This figure shows how interframe generation works. When a linear 
% relationship between the latent space and the time axis is established, it is possible to create a frame at an   arbitrary % time position with only alpha blending.}
% \label{fig:recon}
% \end{figure}

\vspace{-3mm}

\subsection{\sam{Training} method and settings}

The proposed \hdh{training} method has two main types.
The first is the offline training method, which trains the network using the entire training set, and then generates interframes with these pre-trained parameters. %In this case, only one parameter is formed that fits the entire training set, and interframes can be generated simply by loading the pre-trained model at the inference.
The second method is the online \hdh{training} method \hdh{that fine-tunes network} using \hdh{nearby} frames \iffalse that are actually referenced.\fi\hdh{with target frames.} In this case, since additional \iffalse learning\fi\hdh{training} with the inputs must be performed to create each output, \iffalse a lot of\fi\hdh{the training cost is much higher.} 
% However, as online training method have the advantage of learning the surrounding data more accurately, allowing them to better handle \hdh{difficult tasks like} accelerated motions, occluded areas, and complex motions.
\hdh{However, as the online training method enables the network to refer to nearby frames more thoroughly, it allows the network to better handle difficult tasks like accelerated motions, occluded areas, and complex motions.}
In the former case, it can be used when there is a \hdh{limitation} on the cost. In the latter case, it is used when a better result is needed without a cost limit.

The training was performed with the Middlebury dataset\cite{baker2011database}, which is commonly used to see the performance of optical flow.
The Middlebury dataset consists of 11 videos for training and provides the same number of videos for evaluation. Each video consists of 8 consecutive frames, and the optical flow ground truth is provided only in the training set. \hdh{However,} we did not refer to this ground truth at all \hdh{which differentiates our network from other conventional approaches.}

In offline training, \hdh{the network was trained} to reduce $Loss_L$ and $Loss_I$ \hdh{using} consecutive 3 frames \hdh{made} by \hdh{the} frame sliding method. Because the frame-sliding method is adopted, 6 \hdh{sets} of inputs \hdh{per} video, and 66 training data \hdh{were} used \hdh{as the training set in total}. We trained the entire training set \hdh{with} 200 epochs and saw this as the result of offline learning. As a hyperparameter for training, the initial LR was 0.1, LR decays by 0.95 times for each epoch, and %we used SGD as the optimizer. 
\hdh{SGD was used as an optimizer.} \sam{The weights we used for $w_L$ and $w_I$ are 0.1 and 1.0.} \hdh{The $w_I$ is greater because restoring a frame is our primary purpose, not the discriminator.}

Online \hdh{training starts with pre-trained} offline parameters and additional \hdh{training is} performed for each evaluation video. Specifically, since the Middlebury evaluation set checks the \hdh{inferred frame} of the 10$^{th}$ frame, we \hdh{trained the network with} 1500 iterations using two training samples: frame 7-8-9 and 11-12-13. %At this time, the reason we learned only the correlation of the three frames before and after is because we should not refer to frame 10 for a fair evaluation.
%\hdh{For fair evaluation of our novel network, sample sets that contained 10th frame was not included in training.}
\sam{For a fair evaluation of our novel method, the 10$^{th}$ frame was not included in the training.}

\begin{figure*}
%     \begin{tabular}{ccc}
%     Ground truth&Interpolation on latent space&Interpolation on input space
%     \\
%     \includegraphics[width=4cm]{png/000_answer.png}&\
%     \includegraphics[width=4cm]{png/000_mine.png}&
%     \includegraphics[width=4cm]{png/000_ref.png}
%     \\
%     \includegraphics[width=4cm]{png/001_answer.png}&
%     \includegraphics[width=4cm]{png/001_mine.png}&
%     \includegraphics[width=4cm]{png/001_ref.png}
%     \\
% %    \includegraphics[width=4cm]{png/007_answer.png}&
% %    \includegraphics[width=4cm]{png/007_mine.png}&
% %    \includegraphics[width=4cm]{png/007_ref.png}
% %    \\
% %    \includegraphics[width=4cm]{png/009_answer.png}&
% %    \includegraphics[width=4cm]{png/009_mine.png}&
% %    \includegraphics[width=4cm]{png/009_ref.png}
% %    \\
%     \includegraphics[width=4cm]{png/004_answer.png}&
%     \includegraphics[width=4cm]{png/004_mine.png}&
%     \includegraphics[width=4cm]{png/004_ref.png}
%     \end{tabular}
    \begin{tabular}{ccc}
    Ground truth&Interpolation on latent space&Interpolation on input space
    \\
    \includegraphics[width=4cm]{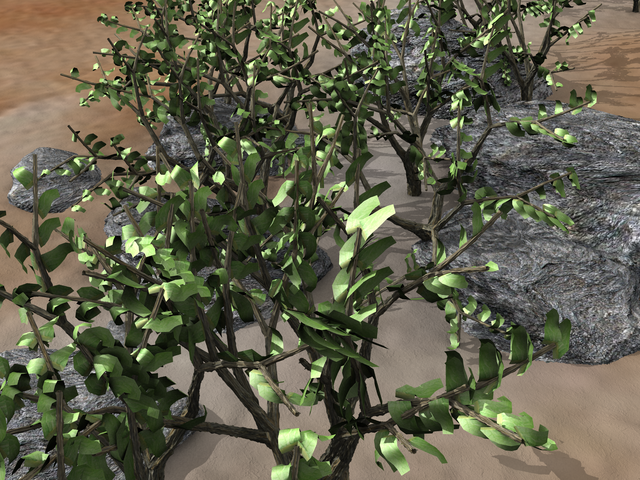}&\
    \includegraphics[width=4cm]{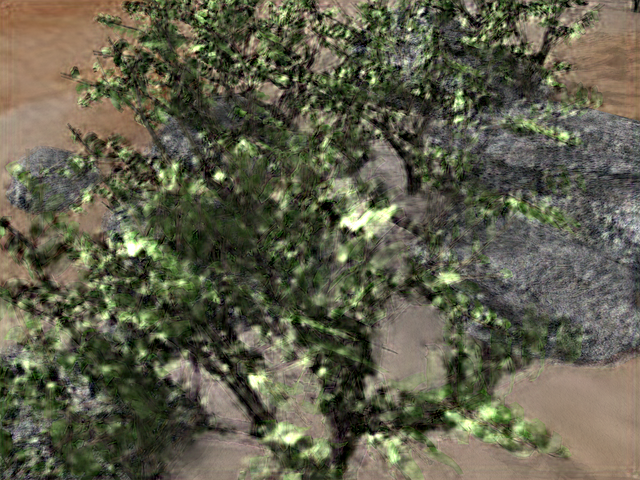}&
    \includegraphics[width=4cm]{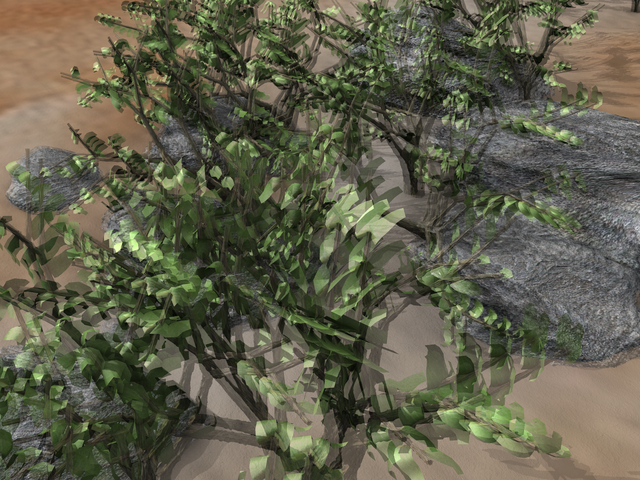}
    \\
    \includegraphics[width=4cm]{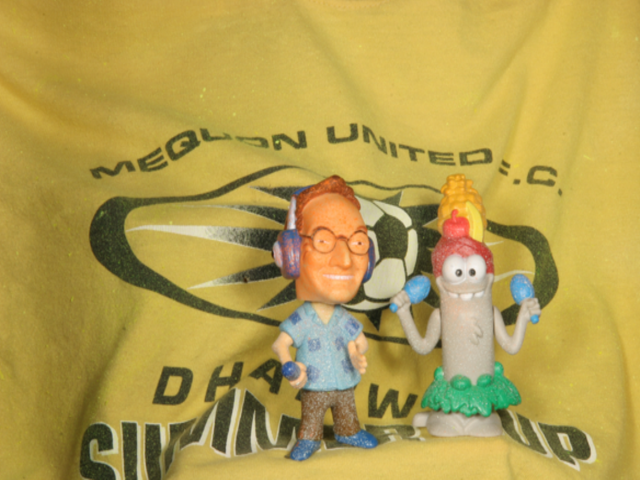}&
    \includegraphics[width=4cm]{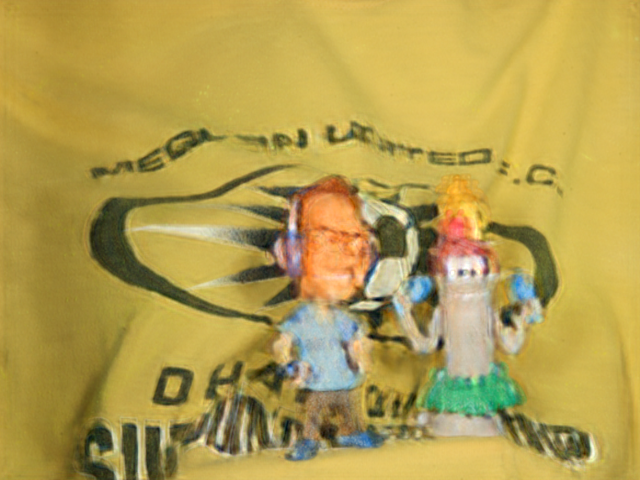}&
    \includegraphics[width=4cm]{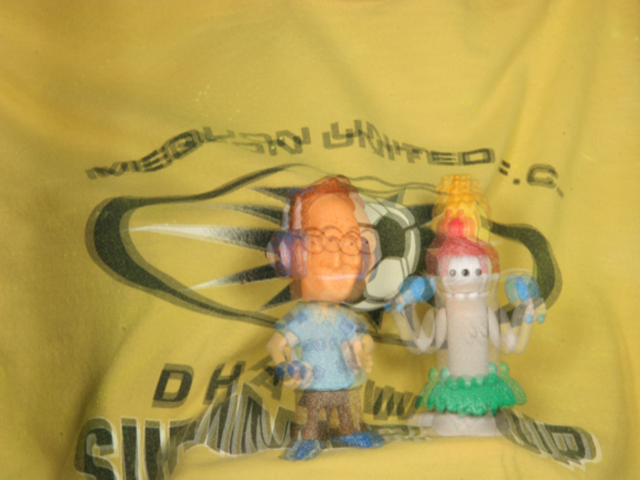}
    \\

    \includegraphics[width=4cm]{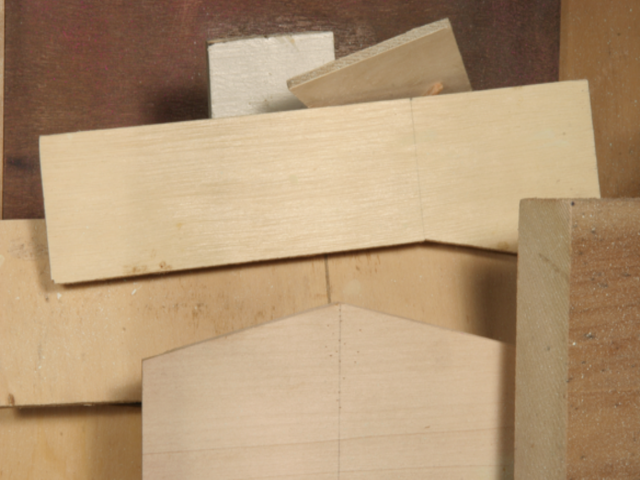}&
    \includegraphics[width=4cm]{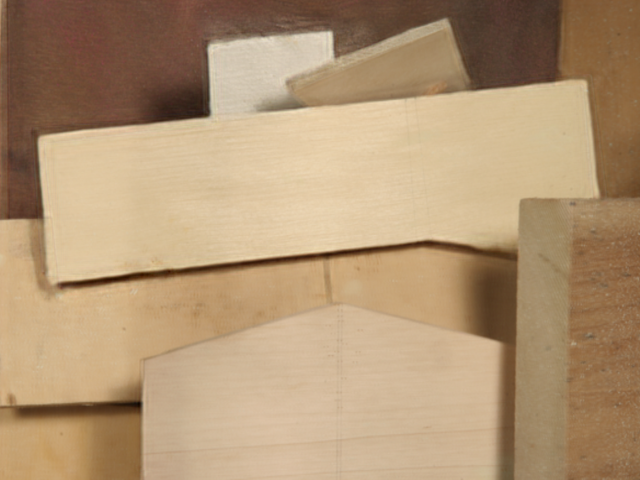}&
    \includegraphics[width=4cm]{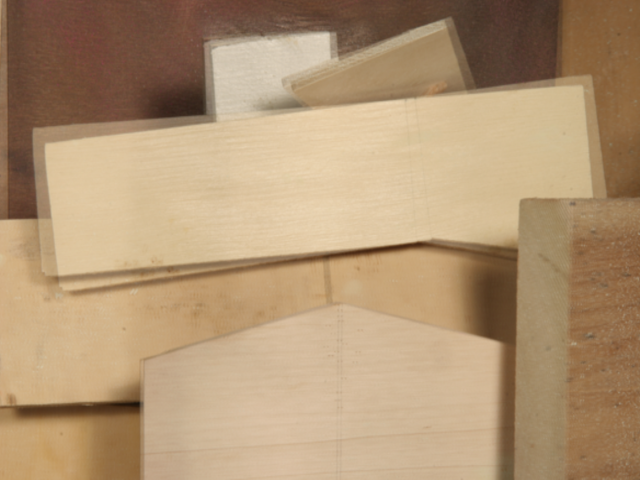}
    \end{tabular}
    \vspace{-5mm}
    \caption{\footnotesize{\textbf{\sam{Interpolation result on latent sapce vs input space}} The middle row is an intermediate frame created with only linear interpolation in the learned latency space of UGLOW without using any optical flow at all. Unlike blending in the input space on the right, the proposed method makes a frame corresponding to the actual middle position.}
    } %We can see that it's not all about mimicking the original.}}
\label{result_in_image_1}
\end{figure*}

\section{Result}

% In this section, we are going to evaluate the performance of the proposed method by generating the 10$^{th}$ frame of the Middlebury evaluation set.

\begin{table}
\caption{% {Table 1과 그림 5가 동일한 정보이므로 table 1을 제거} 
\footnotesize{\textbf{PSNR and SSIM.}  \hspace{3mm} {This table shows the experimental results of PSNR and SSIM, which reproduced frame 10 of the Middlebury evaluation set using only frames 9 and 11. The proposed algorithm shows improved evaluation values compared to simple image blending in all videos.}}
}
\vspace{-2mm}
\begin{center}
\begin{tabular}{|c||c|c||c|c|}
\hline
	Video & PSNR & PSNR & SSIM & SSIM \\
	Name & (ref) & (ours) & (ref) & (ours) \\
\hline\hline
	\texttt{Grove} & 15.902 & 16.771 & 0.2492 & 0.3221\\
	\texttt{Mequon} & 23.220 & 25.056 & 0.7377 & 0.7965\\
	\texttt{Yosemite} & 27.109 & 29.399 & 0.7737 & 0.8450\\
	\texttt{Dumptruck} & 24.698 & 25.019 & 0.9185 & 0.9199\\
	\texttt{Wooden} & 27.156 & 32.433 & 0.8516 & 0.8902\\
	\texttt{Army} & 33.885 & 34.919 & 0.9297 & 0.9323\\
	\texttt{Basketball} & 23.976 & 25.962 & 0.8520 & 0.8760\\
	\texttt{Evergreen} & 23.353 & 24.523 & 0.7783 & 0.8106\\
	\texttt{Backyard} & 22.081 & 23.260 & 0.6877 & 0.7047\\
	\texttt{Schefflera} & 25.549 & 26.805 & 0.6535 & 0.6956\\
	\texttt{Urban} & 23.003 & 25.126 & 0.6004 & 0.6414\\
\hline
	\texttt{Average} & 24.539 & 26.298 & 0.7302 & 0.7668\\
\hline
\end{tabular}

\end{center}
\vspace{-4mm}
\label{table:measure_result}
\end{table}

\subsection{Objective evaluation}
In this section, we are going to evaluate the performance of the proposed method by generating the 10$^{th}$ frame of the Middlebury evaluation set. For an objective comparison, we measured the difference between our result and the \hdh{10$^{th}$ frame from} the Middlebury evaluation set. PSNR and SSIM were used here.
The results show higher values, as shown in the Table\ref{table:measure_result}.

\subsection{\hdh{Empirical} evaluation}
Fig \ref{result_in_image_1} shows some examples of \hdh{our experiment}.
The image on the left is ground truth, and the image in the middle shows the middle frame transformed by linear blending in latent space by entering frames 9 and 11 of the Middlebury evaluation set. The image on the right is the result of performing the same linear interpolation in image space, not in latent space.
In the image on the right, we can see how fast the object is moving between the two frames. In latent space, the proposed method combines two distant objects and complex details.
%Although the quality is lower than the image result created by applying shift by optical flow, it can be said of high value to form the first meaningful intermediate image with pure linear interpolation without optical flow.
\hdh{Although optical flow output may seem more elaborate, our U-Net based Generative Flow has its originality in the methodology. 
The U-Net based Generative Flow only uses linear interpolation, unlike other conventional approaches that rely on optical flow.}
%Another great advantage is that  the parts of high difficulties, such as the face of the doll, the background of the leaves, and the texture of the stones, are similar to the result of simple blending without artifacts. This helps produce robust results when processing video.
\hdh{Another great advantage is that our outputs showed similarities with the outputs made with simple blending on difficult tasks such as the face of the doll, leaves from the background, and the texture of the stones.}
%This helps guarantee stable frames when processing video.

\section{Conclusion}
In this paper, we proposed a new method of generating intermediate frames using video data itself without making optical flow information using an invertible deep neural network. We proposed UGLOW, a reversible network that produces better results, and confirmed its feasibility using the Middlebury data set.
We developed a loss that induces a temporal linear relationship between successive frames of video in a latent space and proposed an algorithm capable of generating mid-view results using a trained reversible network. %We have shown that this intuitively made method has a meaningful effect through image results and objective results. 
\hdh{We have shown that this intuitive approach made plausible results through empirical and objective measures.}

The \hdh{biggest} \iffalse advantage\fi\hdh{contribution} of our proposal is that it is the first attempt \iffalse to\fi not \hdh{to} use optical flow for video interpolation.
%This can be seen as a good attempt in that deep learning learns everything by itself without forcing it to learn, and it is considered a suitable direction for video processing in the future.
\hdh{This aligns with the paradigm that deep learning can learn everything without relying on a knowledge-based system.}
As future works, we will verify this proposal in various test sets and improve the performance to be similar to the model using optical flow.

\newpage

\bibliographystyle{splncs04}
\bibliography{egbib.bib}

\end{document}